\documentclass[10pt,twocolumn,letterpaper]{article}

\usepackage[pagenumbers]{cvpr} 

\definecolor{cvprblue}{rgb}{0.21,0.49,0.74}
\usepackage[pagebackref,breaklinks,colorlinks,allcolors=cvprblue]{hyperref}
\usepackage{url}

\usepackage{booktabs}
\usepackage{multirow}
\usepackage{colortbl}
\usepackage{xcolor}
\usepackage{graphicx} 
\usepackage{amsmath}
\usepackage{amssymb}
\usepackage[ruled,vlined]{algorithm2e}
\usepackage{pifont}
\def\paperID{9771} 
\def\confName{CVPR}
\def\confYear{2026}

\title{Revisiting Image Manipulation Localization under Realistic Manipulation Scenarios}

\author{
Xuekang Zhu$^{1,2\dagger}$ \quad
Ji-Zhe Zhou$^{1\dagger\ddagger}$ \quad
Kaiwen Feng$^{1}$ \quad
Chenfan Qu$^{3,2}$ \quad
Xiwen Wang$^{1}$ \\ 
Yunfei Wang$^{1}$ \quad 
Liting Zhou$^{1}$  \quad
Jian Liu$^{2\ddagger}$ \\
$^{1}$Sichuan University \quad
$^{2}$Ant Group \quad
$^{3}$South China University of Technology
}

\begin{document}
\maketitle

\renewcommand{\thefootnote}{\dag}
\footnotetext{Equal contribution. Corresponding authors: Jian Liu (\href{mailto:rex.lj@antgroup.com}{rex.lj@antgroup.com}) and Jizhe Zhou (\href{mailto:jzzhou@scu.edu.cn}{jzzhou@scu.edu.cn}).}

\begin{abstract}
With the large models easing the labor-intensive manipulation process, image manipulations in today’s real scenarios often entail a complex manipulation process, comprising a series of editing operations to create a deceptive image. However, existing IML methods remain manipulation-process-agnostic, directly producing localization masks in a one-shot prediction paradigm without modeling the underlying editing steps. This one-shot paradigm compresses the high-dimensional compositional space into a single binary mask, inducing severe \textbf{dimensional collapse}, which forces the model to discard essential structural cues and ultimately leads to overfitting and degraded generalization. To address this, we are the first to reformulate image manipulation localization as a conditional sequence prediction task, proposing the RITA framework. RITA predicts manipulated regions layer-by-layer in an ordered manner, using each step's prediction as the condition for the next, thereby explicitly modeling temporal dependencies and hierarchical structures among editing operations. To enable training and evaluation, we synthesize multi-step manipulation data and construct a new benchmark HSIM. We further propose the HSS metric to assess sequential order and hierarchical alignment. Extensive experiments show that: 1) RITA achieves SOTA generalization and robustness on traditional benchmarks; 2) it remains computationally efficient despite explicitly modeling multi-step sequences; and 3) it establishes a viable foundation for hierarchical, process-aware manipulation localization. Code and dataset are available at \url{https://github.com/scu-zjz/RITA}.
\end{abstract}
\section{Introduction}
Creating deceptive images typically involves a series of multi-step, even nested manipulation operations to produce fine details and hide human-perceptible artifacts. With the aid of large models, the labor-intensive and professional-skill-intensive manual manipulation process has been significantly simplified into a semi-automatic manner, requiring only simple prompt engineering~\cite{sun2024rethinking,wang2025opensdi}. As a result, deceptive images involving complex manipulation processes are now ubiquitous in real life~\cite{NIST16_2019}. Revealing such deceptive images has thus become an urgent need in the field of Image Manipulation Localization (IML).
\begin{figure}[t]
    \centering
    \includegraphics[width=\linewidth]{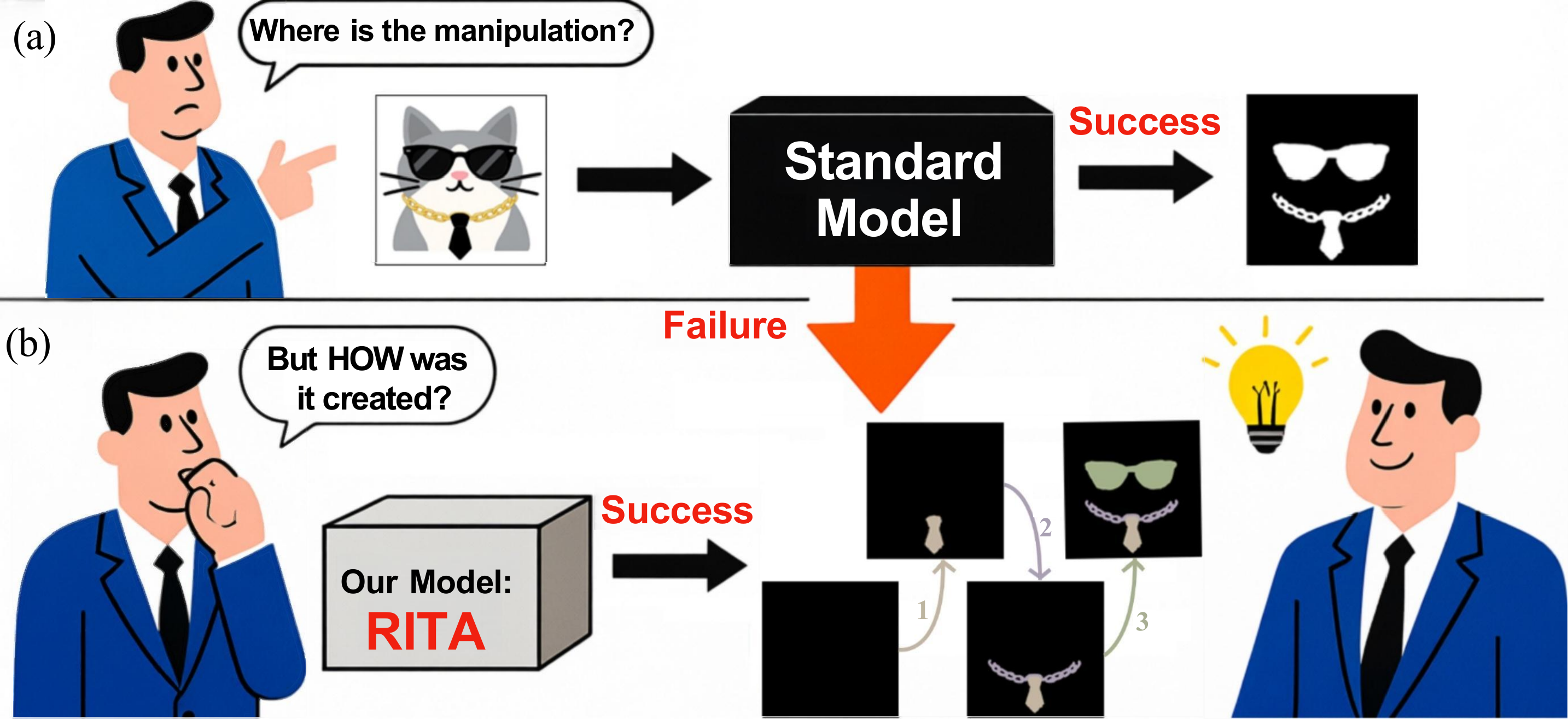}
    \caption{
        Comparison between (a) the standard one-shot localization and (b) our RITA framework. 
    }
    \label{fig:intro}
\end{figure}
However, when applied to real scenarios, existing IML methods often experience a significant performance drop~\cite{luo2024toward}. Most existing studies~\cite{park2025community} attribute this to training data bias and have made various efforts to bridge the data distribution gap.

Considering the manipulation processes involved in generating deceptive images, we identify a long-neglected yet pivotal factor behind this performance gap: \textbf{dimensional collapse} induced by the one-shot prediction paradigm.

Most existing IML methods~\cite{MVSS_2021,objectformer_2022,CAT-Net2022,liu2022pscc,trufor2023,zhu2025mesorch,tan2025veritas} adopt a one-shot prediction paradigm. In this paradigm, a model inputs an image and directly outputs a unified binary mask (Fig.~\ref{fig:intro}(a)) to identify all potentially manipulated regions, answering \textit{where} the manipulation occurred. This paradigm indeed offers enhanced interpretability over simple binary classification and has achieved notable success on current benchmarks. Nevertheless, it overly simplifies the multi-stage synthesis pipelines used in existing datasets.

In fact, although manipulated images in existing IML datasets~\cite{CASIA_2013,NIST16_2019,IMD20_2020,trufor2023,jia2023autosplice} are generated through a sequence of editing operations, the annotation only provides the final binary mask while discarding all intermediate steps.
For example, in AutoSplice~\cite{jia2023autosplice}, a manipulated sample is produced through several explicit stages: selecting and erasing an object mask, compositing a semantically matched region from another image, and performing boundary harmonization via Poisson or Laplacian blending. Such a process exhibits two distinct characteristics: 1) Sequentiality, as edits must be applied in a specific temporal order, and 2) Hierarchy, as later operations often build upon prior edits, forming spatial and semantic dependencies. Thus, the prevailing one-shot paradigm in IML only outputs a single mask that answers \textit{Where} manipulation occurred, collapsing the high-dimensional compositional space of sequential and hierarchical edits into a single binary mask. This \textbf{dimensional collapse} forces the model to discard crucial temporal and structural cues, leaving it to learn only coarse spatial distributions. As a consequence, its ability to generalize to diverse and unseen manipulations is inherently limited.


To resolve this problem, we are the first to reformulate the task of image manipulation localization from a one-shot prediction problem to a \textbf{conditional sequence prediction} problem. Accordingly, we design a novel autoregressive localization framework, termed \textbf{RITA} (Reversely-ordered Incremental-Transition Autoregression). As illustrated in Fig.~\ref{fig:intro}(b), the core idea of RITA is to progressively and explicitly answer \textit{how} a manipulation is constructed by predicting manipulation regions layer by layer in an ordered sequence, conditioning each prediction on the input image and the output from the immediately preceding step. This design explicitly models the temporal dependencies and hierarchical structures among editing steps, thereby naturally and effectively disentangling the layered composition of multi-step manipulations and resolving the dimensional mismatch inherent in the data.

To enable training and evaluation for this new paradigm, we design a tree-structured reverse sampling strategy to construct a synthetic dataset that simulates the multi-step hierarchical editing process. We also annotate the Hierarchical Sequential Image Manipulation (HSIM) dataset, the first benchmark that provides explicit, stepwise annotations for multi-step manipulation trajectories. Furthermore, we introduce the {Hierarchical Sequential Score (HSS), a metric that measures the accuracy of predicted sequences in terms of both sequential order and hierarchical alignment.

At the same time, our paradigm remains compatible with traditional one-shot datasets by treating them as a special case of a two-step process. Extensive experiments demonstrate that: 1) Compared with existing IML baselines, our sequence modeling approach achieves stronger cross-dataset generalization and improved robustness to diverse manipulation types; 2) Despite explicitly modeling multi-step sequences, RITA maintains competitive parameter count, computational cost, and inference efficiency; 3) Together with our synthetic multi-step data and the HSIM benchmark, this process-centric formulation validates the feasibility of temporal sequence prediction for IML. It also represents an important first step toward downstream tasks that require reasoning about intermediate editing stages rather than only final outcomes.

Our main contributions are summarized as follows:
\begin{itemize}
    \item We reformulate manipulation localization as a sequence prediction problem and introduce RITA, the first framework that explicitly models the temporal and hierarchical structure of editing operations, enabling process-centric localization.

    \item We build HSIM, the first dataset with stepwise multi-stage annotations, and design a tree-based sampling strategy to generate hierarchical synthetic trajectories. We further propose HSS, a metric that evaluates both structural and sequential consistency.

    \item Extensive experiments show that RITA delivers stronger generalization and robustness, preserves competitive efficiency despite multi-step modeling, and validates temporal sequence prediction as a viable foundation for future process-centric manipulation tasks.
\end{itemize}

\section{Related Works}
Most IML methods~\cite{Mantra_2019,MVSS_2021,CAT-Net2022,trufor2023,zhu2025mesorch,liu2022pscc,Qu_2023_CVPR,ma2023iml,qu2024omni,qu2024towards,su2025spase,qu2025revisiting,qu2026textshield,tan2026videoveritas} follow a unified paradigm: given an image, the model outputs a single binary mask covering all manipulated regions. This formulation collapses the inherently multi-dimensional nature of manipulation into a single-step prediction, ignoring the sequential and structural aspects of the editing process. SAFIRE~\cite{kwon2025safire} emphasizes source separation, but similarly overlooks temporal dependencies within edits from the same source. In effect, multi-step manipulations are still reduced to a flattened representation, indicating that source-level cues alone are insufficient to capture the full dimensionality of manipulation required for fine-grained analysis.

Many widely used IML datasets~\cite{CASIA_2013,Columbia_2006,NIST16_2019,IMD20_2020,trufor2023,jia2023autosplice} are generated through predefined editing pipelines. In these pipelines, forged images are typically produced by a sequence of operations such as object erasing, inpainting, and boundary blending. However, only the final manipulation mask is retained in the annotation, while intermediate stages are discarded. This design obscures the underlying multi-step nature of the manipulation process and forces models to learn from a flattened, single-stage supervision signal, making it difficult for them to capture how complex forgeries are actually constructed. On the evaluation side, recent benchmarks adopt the Binary F1 score introduced by IMDL-BenCo~\cite{ma2025imdl} as the primary localization metric, which further focuses assessment on the final mask and remains agnostic to errors in the predicted edit sequence.

\begin{figure*}[t]
    \centering
    \includegraphics[width=0.995\textwidth]{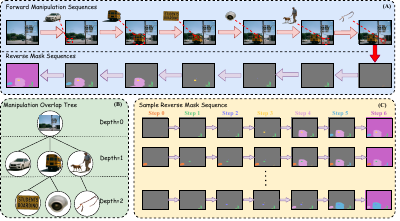}
    \caption{Illustration of the synthetic multi-step manipulated process. 
(A) Sequential application of localized manipulations to an image, which depicts both the macroscopic manipulated process and the prediction of a corresponding mask at each step. 
(B) Organization of the manipulated regions into a hierarchical tree structure, where the figure shows the constructed tree for the current example. 
(C) Reverse-order sampling from the manipulation tree to construct mask sequences, where manipulated areas are accumulated step by step. 
Gray regions denote padding used to align the mask sequences.}
    \label{fig:tampering_process}
\end{figure*}

\section{Method}

\paragraph{Problem Formulation: Conditional Sequence Prediction.} We reformulate manipulation localization as a conditional sequence prediction task, progressively predicting a sequence of masks $(M_1, \dots, M_T)$ to reveal the manipulation history. While the space of all mask sequences is the vast Cartesian product $\mathcal{M}^T$ (where $\mathcal{M}=\{0,1\}^{H \times W}$), valid sequences must obey the \textbf{progressive containment property}:
\begin{equation}
M_t \subseteq M_{t+1} \quad \text{for all } t \in \{1, \dots, T-1\}.
\end{equation}
This constraint dictates that manipulated regions only accumulate, meaning our goal is to predict a monotonically increasing path of masks. As shown in Figure~\ref{fig:tampering_process}, our approach is built upon three key steps:
(A) sequential localized manipulations to form composite manipulations, 
(B) hierarchical organization of manipulated regions into a tree structure, 
and (C) reverse-order sampling of manipulation masks from the tree to form autoregressive training sequences.

We first  describe dataset construction and evaluation metrics (Sec.~\ref{subsec:data_and_metrics}), 
formalize the conditional Markov process underlying conditional sequence prediction (Sec.~\ref{subsec:markov}),
and present our model design (Sec.~\ref{subsec:model_design}).

\subsection{Dataset Construction and Evaluation Metrics}
\label{subsec:data_and_metrics}

To enable fair evaluation on both traditional benchmarks and multi-step sequence scenarios, we unify all datasets under a sequence prediction paradigm. Specifically, for traditional one-shot datasets such as CASIAv2~\cite{CASIA_2013}, we reformulate them into two autoregressive steps, while we also construct new datasets with multiple manipulated operations to reflect multi-step manipulation paths. We also introduce a new metric to evaluate this new scenario. This unified design allows consistent training and evaluation across both traditional and hierarchical manipulation cases.

\subsubsection{Sequence Manipulated Dataset Construction}
\label{subsubsec:tree_data}
To train and evaluate our model on hierarchical manipulations, we synthesize a sequence manipulation dataset by extending benchmarks like CASIAv2. We apply a series of random copy-move operations, where each new region corresponds to a \textbf{node} in a tree. The spatial relationships between these operations, such as nesting and adjacency, naturally form a hierarchical structure. This hierarchy is formally captured as a \textbf{manipulation tree}, which serves as the multi-step representation for the manipulation process.

\paragraph{Tree-Structured Manipulation Representation.} As illustrated in Fig.~\ref{fig:tampering_process}(B), we represent the spatial hierarchy of manipulated regions as a tree $\mathcal{T}=(V, E)$, where each node $v \in V$ corresponds to a distinct manipulated region $R(v)$. The tree is constructed dynamically based on a clear hierarchical rule: the parent of a new region is the deepest node in the existing tree that contains this new region.
\begin{equation}
\text{parent}(v_j) = \underset{v \in V' \text{ s.t. } R(v_j) \subseteq R(v)}{\text{argmax}} \left( \text{depth}(v), \text{Area}(R(v)) \right).
\end{equation}
Here, $V'$ denotes the set of pre-existing nodes. An edge $(v_i \to v_j)$ is then added to $E$, defining the containment hierarchy. The root node $v_{\text{root}}$ represents the full image domain, and the tree correctly simulates the partial order of operations.

\paragraph{Sequential Path Sampling.}
To capture this diversity for training, we generate $N$ distinct sequences by performing reverse-order sampling from the manipulation tree (Fig.~\ref{fig:tampering_process}(C)), which emulates different plausible editing orders without altering the underlying hierarchy. Each path, denoted $\mathcal{P}_i = (v_1, \dots, v_{T_i})$, is constructed via \textbf{reverse-sampling}—a process of iteratively selecting and removing a random leaf node from the tree. At each step $k$, a node is drawn uniformly from the current set of leaf nodes:
\begin{equation}
v_k \sim \text{Uniform}(\text{Leaves}(\mathcal{T} \setminus \{v_1, \dots, v_{k-1}\})).
\end{equation}
From this path, we construct the final accumulated mask sequence $\{M_t^{(i)}\}$. The mask at each step $t$ is the union of the regions from the first $t$ nodes in the path:
\begin{equation}
M_t^{(i)} = \mathrm{Union}_{j=1}^{t} R(v_j).
\end{equation}
This node-centric construction guarantees the progressive containment property ($M_t^{(i)} \subseteq M_{t+1}^{(i)}$), which is essential for autoregressive modeling. It implies that at any step $t$, the model operates on the revealed content within $M_t^{(i)}$, while the unrevealed areas are masked out using padding. Even for regions without spatial inclusion, path sampling still provides valid temporal orders, allowing the model to learn multi-step sequences without requiring hierarchy.

\paragraph{Hierarchical Sequential Image Manipulation Dataset.}
\label{sec:hsim}
To evaluate under multi-step scenario, we construct a synthetic dataset. Starting from a pool of authentic photographs, we generate stepwise manipulation plans with GPT-4o and execute the edits using GPT-Image-1. AI editing often introduces \emph{undesired artifacts}, where regions outside the intended manipulated area are inadvertently modified or distorted~\cite{costanzino2025towards}. To mitigate this, we perform targeted manual refinements so that only the intended regions are altered while the surrounding content remains intact. After sampling, HSIM contains \textbf{2,442 valid manipulation paths}, each supplied with stepwise masks for autoregressive evaluation.
\begin{figure*}[t]
    \centering
    \includegraphics[width=0.95\textwidth]{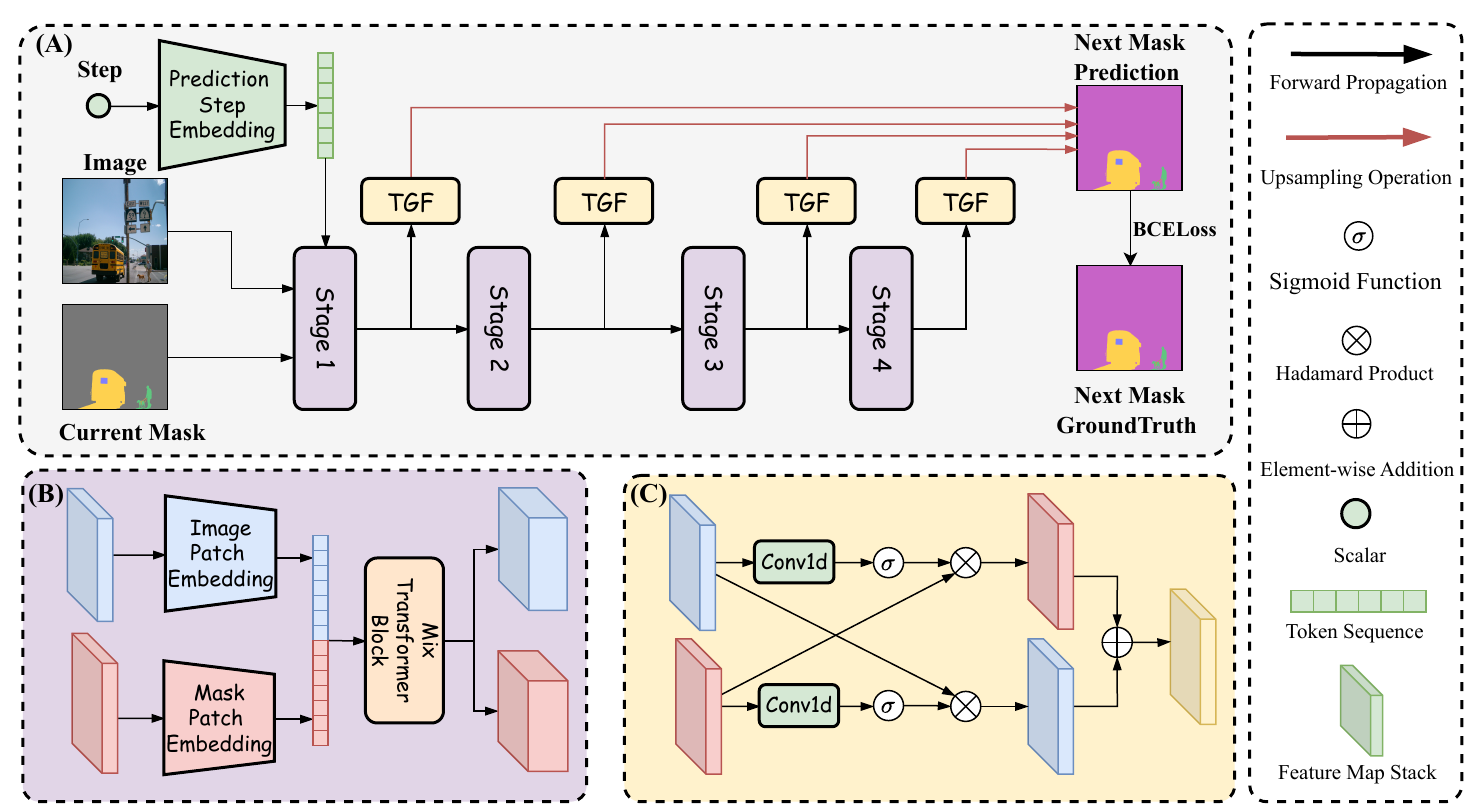}
    \caption{Overall architecture of our proposed framework.(A) Autoregressive prediction pipeline that encodes the image, current mask, and step embedding through multi-scale stages with TGF modules.(B) Multi-scale encoder that jointly processes image and mask tokens via a Mix Transformer block.(C) Transition Gated Fusion module that gates image and mask features to evolving manipulated regions.}
    \label{fig:framework}
\end{figure*}

\subsubsection{Compatibility with Existing Datasets}
\label{subsec:compatibility}

For existing datasets such as CASIAv2~\cite{CASIA_2013}, which only provide a single-step binary mask, 
we design a simple yet effective decomposition strategy to make them compatible with our autoregressive paradigm. 
Specifically, the image is split into two regions: the manipulated region $M_{\text{tamper}}$ 
and the authentic region $M_{\text{auth}} = 1 - M_{\text{tamper}}$. 
This enables us to reinterpret the binary annotation as a \textbf{two-step} incremental sequence.

The mask sequence is constructed as:
\begin{equation}
M_t = 
\begin{cases}
M_{\text{tamper}} \cup M_{padding}, & t = 1, \\
M_{\text{full}}, & t = 2,
\end{cases}
\label{eq:single_step_decomp}
\end{equation}
where $M_{padding}$ denotes a placeholder region for alignment.

At $t=1$, the model focuses solely on the manipulated region, while the authentic region is padded out. 
At $t=2$, the model predicts the full image domain, including both manipulated and authentic regions. 
This design ensures that one-shot datasets can be seamlessly incorporated into our sequence autoregressive paradigm, 
maintaining consistency across all training and evaluation settings.

\subsubsection{Hierarchical Sequential Score}
\label{sec:hss}
To evaluate predictions on our tree-structured data, we must account for the non-unique manipulation orders. We therefore evaluate a single prediction against all $N$ valid paths that can be derived from a manipulation tree $\mathcal{T}$ (Fig.~\ref{fig:tampering_process}(C)), combining structural alignment and length consistency.

\paragraph{Structure Matching.}
To handle discrepancies in length between predicted sequences and ground-truth paths, we utilize MonotonicMatch, a dynamic programming algorithm (detailed in the Supplementary Material). For each ground-truth path, this algorithm computes an optimal monotonic, non-decreasing alignment with the predicted sequence, maximizing the stepwise F1 score between them.

Let the predicted mask sequence be $P \in \mathbb{R}^{T_p \times H \times W}$ with $T_p$ steps, and the ground-truth path set be $M \in \mathbb{R}^{N \times T_g \times H \times W}$ containing $N$ sampled paths each with $T_g$ steps. For each path $M_i$, we compute
\begin{equation}
\text{F1}_{\text{match}} = \max_{i=1,\dots,N} \text{MonotonicMatch}(P, M_i),
\end{equation}

\paragraph{Final Score with Length Consistency.}

To penalize mismatched step counts between prediction length $T_p$ and ground-truth length $T_g$, we apply a length penalty:
\begin{equation}
\lambda(T_p, T_g) = \exp\!\left(-\alpha \frac{(T_p - T_g)^2}{\max(T_g,1)}\right),
\end{equation}
where $\alpha=0.1$ controls the penalty strength. The final score is defined as
\begin{equation}
\text{Score} = \lambda(T_p, T_g) \times \text{F1}_{\text{match}}.
\end{equation}
This metric, termed the \textbf{Hierarchical Sequential Score (HSS)}, jointly evaluates sequential accuracy and structural alignment, ensuring fair assessment across multiple valid manipulation paths.

\subsection{Conditional Markov Formulation}
\label{subsec:markov}
We formulate mask prediction as a conditional sequence prediction task. In general, predicting the next mask $M_{t+1}$ would depend on the input image $I$ and the all history of previous masks $(M_0, \dots, M_t)$. However, due to the \textbf{progressive containment property} ($M_t \subseteq M_{t+1}$), the current mask $M_t$ acts as a sufficient statistic, encapsulating all spatial information from the preceding steps. This insight allows us to simplify the dependency to a \textbf{first-order Markov assumption}:
\begin{equation}\label{eq:markov}
P(M_{t+1} \mid I, M_t, \dots, M_0) = P(M_{t+1} \mid I, M_t).
\end{equation}
Consequently, the full sequence probability decomposes autoregressively, reducing our task to learning the one-step transition model $P(M_{t+1} \mid I, M_t)$:
\begin{equation}
P(M_1, \dots, M_T \mid I) = \prod_{t=0}^{T-1} P(M_{t+1} \mid I, M_t).
\end{equation}

\subsection{Model Design}
\label{subsec:model_design}

Our framework follows a conditional first-order Markov structure~\ref{subsec:markov}, where the prediction of the next mask $M_{t+1}$ depends on the current mask $M_t$ and the input image $I$.

\paragraph{Architecture Overview.}  
As illustrated in Figure~\ref{fig:framework}(A), the model takes as input the manipulated image $I$, the current mask $M_t$, and the step index $s_t$. A learnable embedding maps $s_t$ to a vector $E_t$, which is injected into the first-stage mask features to provide temporal conditioning. The inputs are processed by a multi-scale encoder, transition-gate fusion modules, and a decoder that outputs the next mask prediction $M_{t+1}$.

\paragraph{Multi-Scale Encoder.}  
The encoder design is shown in Figure~\ref{fig:framework}(B). The image and mask are separately processed into hierarchical feature maps:
\begin{equation}
\{F_I^{(l)}\}_{l=1}^4 = \text{Stage}_I(I), 
\quad 
\{F_M^{(l)}\}_{l=1}^4 = \text{Stage}_M(M_t),
\end{equation}
where $l$ denotes the stage index. At the first stage, the mask features are augmented with the broadcasted step embedding:
\begin{equation}
F_M^{(1)} \leftarrow F_M^{(1)} + E_t.
\end{equation}
At each stage, image and mask tokens are concatenated, passed through a Mix Transformer block, and split again into two streams, enabling joint reasoning while preserving modality-specific information.

\paragraph{Transition Gated Fusion.}  
To capture manipulation cues, we introduce a Transition Gated Fusion(TGF) module at each scale (Figure~\ref{fig:framework}(C)). Given image and mask features, we compute:
\begin{align}
G_I^{(l)} &= \sigma(W_M F_M^{(l)}), \\
G_M^{(l)} &= \sigma(W_I F_I^{(l)}), \\
\hat{F}^{(l)} &= F_I^{(l)} \odot G_M^{(l)} + F_M^{(l)} \odot G_I^{(l)},
\end{align}
where $W_I$ and $W_M$ are $1{\times}1$ convolutions, $\sigma(\cdot)$ is the sigmoid function, and $\odot$ denotes element-wise multiplication. This transition-gating mechanism ensures that image features are modulated by mask context and vice versa, focusing attention on evolving manipulated regions.

\paragraph{Decoder and Mask Prediction.}  
Fused features from all scales are upsampled and concatenated with mask features:
\begin{equation}
\begin{aligned}
Z_t = \text{Concat}(
&\hat{F}^{(1)}, \hat{F}^{(2)\uparrow}, \hat{F}^{(3)\uparrow\uparrow}, \hat{F}^{(4)\uparrow\uparrow\uparrow}, \\
&F_M^{(1)}, F_M^{(2)\uparrow}, F_M^{(3)\uparrow\uparrow}, F_M^{(4)\uparrow\uparrow\uparrow})\uparrow\uparrow,
\end{aligned}
\end{equation}
where $\uparrow$ indicates progressive upsampling by a factor of $2$ at each stage. A $1{\times}1$ convolutional classifier then outputs per-pixel class probabilities:
\begin{equation}
p_\theta^{(t)} = \text{Softmax}(W_{\text{cls}} * Z_t).
\end{equation}

\paragraph{Training Objective.}  
We train the model autoregressively with \textbf{teacher forcing}. 
Initialization uses $M_0$, an all-padding mask denoted \texttt{START}. 
At each step $t$, the model predicts $M_{t+1}$ conditioned on $(I, M_t^{\text{GT}})$ and is supervised by per-pixel cross-entropy against $M_{t+1}^{\text{GT}}$. 
The ground-truth sequence ends with an \texttt{EOS} mask, which the model learns to emit as the stop signal. 
The loss is defined as:

\begin{equation}
\mathcal{L}_{\text{CE}}^{(t)} = -\frac{1}{HW}\sum_{x,y}\log p_\theta^{(t)}[M_{t+1}^{\text{GT}}(x,y)],
\end{equation}
where $p_\theta^{(t)}[\cdot]$ is the predicted probability for the ground-truth class at pixel $(x,y)$.

\paragraph{Edge Supervision Loss.}  
To refine boundary localization, we add an auxiliary edge loss. The incremental manipulated region between two consecutive ground-truth masks is
\begin{equation}
\Delta M_{t+1} = M_{t+1}^{\text{GT}} - M_t^{\text{GT}}.
\end{equation}
Its boundary $\text{Boundary}(\Delta M_{t+1})$ is extracted using morphological dilation. The predicted mask $M_{t+1}^{\text{Pred}}$ is supervised only along these boundary pixels:
\begin{equation}
\mathcal{L}_{\text{Edge}}^{(t)}
= \text{BCE}\!\left(\text{Boundary}(M_{t+1}^{\text{Pred}}), \ \text{Boundary}(\Delta M_{t+1})\right).
\end{equation}
The total training loss at step $t$ is
\begin{equation}
\mathcal{L}^{(t)} = \mathcal{L}_{\text{NMP}}^{(t)} + \beta \,\mathcal{L}_{\text{Edge}}^{(t)},
\end{equation}
with $\beta$ balancing next-mask prediction and edge refinement.

\paragraph{Inference.}  
At inference, the process begins with $M_0$ initialized as a special \texttt{START} mask. At each step $t$, the model predicts a distribution
\begin{equation}
p_\theta^{(t)}(x,y) = f_\theta(I, M_t)(x,y),
\end{equation}
and derives the next mask as
\begin{equation}
M_{t+1}(x,y) = \arg\max_{k} \; p_\theta^{(t)}(x,y,k),
\end{equation}
where $k \in \{0,\dots,K{-}1\}$ includes all semantic classes and a special \texttt{EOS} class. 
The autoregressive process terminates once the proportion of pixels predicted as \texttt{EOS} exceeds a predefined threshold $\tau$ (set to 95\% in our experiments), 
or when the maximum step count $T_{\text{max}}$ is reached.

\section{Experiments}
\begin{table*}[ht]
\centering
\caption{Comparison of F1 scores across datasets. 
The table contains two sections: the upper follows the \textbf{MVSS protocol} and the lower the \textbf{CAT-Net protocol}. Column-wise best scores are in \textcolor{red}{red}, and second-best results are \underline{underlined}.}
\label{tab:performance}
\resizebox{\textwidth}{!}{%
\begin{tabular}{lccccccccc}
\toprule
\multirow{2}{*}{\textbf{Model}} & \multicolumn{1}{c}{\textbf{Source-Aligned}} & \multicolumn{7}{c}{\textbf{Cross-Source}}& \multirow{2}{*}{\textbf{Overall Avg}} \\
\cmidrule(lr){2-2} \cmidrule(lr){3-9}
 & CASIAv1 & Coverage & Columbia & NIST16 & IMD2020 & CocoGlide & Autosplice & \textbf{Cross-Source Avg}  &   \\
\midrule
MVSS-Net & 0.534 & 0.259 & 0.386 & 0.246 & 0.279 & 0.291 & 0.294 & 0.293 & 0.327 \\
CAT-Net  & 0.581 & 0.296 & 0.584 & 0.269 & 0.273 & \underline{0.290} & 0.354 & 0.361 & 0.378 \\
PSCC-Net & 0.378 & 0.231 & 0.604 & 0.214 & 0.235 & 0.227 & \underline{0.652} & 0.377 & 0.363 \\
TruFor   & \underline{0.721} & \underline{0.419} & \textcolor{red}{0.865} & 0.324 & \underline{0.322} & 0.205 & 0.393 & \underline{0.421} & \underline{0.464} \\
Mesorch  & \textcolor{red}{0.740} & 0.326 & 0.726 & \underline{0.343} & 0.269 & 0.162 & 0.249 & 0.346 & 0.402 \\
RITA(Ours)     & 0.522 & \textcolor{red}{0.443} & \underline{0.756} & \textcolor{red}{0.346} & \textcolor{red}{0.379} & \textcolor{red}{0.492} &  \textcolor{red}{0.664} & \textcolor{red}{0.514} & \textcolor{red}{0.515} \\
\bottomrule
MVSS    & 0.583 & 0.482 & 0.740 & 0.336 & -- & 0.443 & 0.385 & 0.477 & 0.495 \\
CAT-Net & 0.808 & 0.427 & \underline{0.915} & 0.252 & -- & 0.410 & 0.387 & 0.478 & 0.533 \\
PSCC    & 0.630 & 0.448 & 0.884 & 0.346 & -- & \underline{0.474} & \underline{0.551} & 0.541 & 0.555 \\
Trufor  & \underline{0.818} & 0.457 & 0.885 & 0.348 & -- & 0.283 & 0.393 & 0.473 & 0.531 \\
Mesorch & \textcolor{red}{0.840} & \textcolor{red}{0.586} & 0.890 & \underline{0.392} & -- & 0.450 & 0.402 & \underline{0.544} & \underline{0.593} \\
RITA(Ours)    & 0.770 & \underline{0.566} & \textcolor{red}{0.921} & \textcolor{red}{0.428} & -- & \textcolor{red}{0.533} & \textcolor{red}{0.643} & \textcolor{red}{0.618} & \textcolor{red}{0.643} \\
\bottomrule
\end{tabular}%
}
\end{table*}

In this section, we report experiments that (1) demonstrate compatibility with the existing one-shot IML paradigm and (2) evaluate performance under newly constructed sequence manipulation scenarios. We further include focused ablations and backbone-controlled comparisons to substantiate generalization and the contribution of each component.

\subsection{Experiments Setup}
All experiments were conducted on eight NVIDIA RTX 3090 GPUs using PyTorch 2.6 with CUDA version 12.4. Input images were uniformly resized to a resolution of 512×512. During training, a batch size of 8 was used, while a batch size of 2 was employed for evaluation. The learning rate followed a cosine decay schedule~\cite{cosine_decay_2017}, starting at 1e-4 and decreasing to a minimum of 5e-7. We adopted the AdamW optimizer~\cite{AdamW_2019} with a weight decay of 0.05 to alleviate overfitting.
\subsection{Evaluation Metric}
For the existing manipulation scenarios, we follow the training and evaluation protocol of the fully aligned IMDLbenco~\cite{ma2025imdl}, using the binary F1 score as the performance metric. Meanwhile, for the sequence manipulation scenario, we adopt the previously introduced metric~\ref{sec:hss} to assess the model's performance under multi-step manipulation scenario.

\subsection{Existing Manipulation Scenarios}
\subsubsection{Performance Comparison}
In this subsection, following Subsection~\ref{subsec:compatibility}, we adapt existing one-shot manipulation datasets into a \textbf{two-step incremental sequence} for our RITA framework. In contrast, existing baselines such as MVSS-Net~\cite{MVSS_2021}, PSCC-Net~\cite{liu2022pscc}, Cat-Net~\cite{CAT-Net2022}, TruFor~\cite{trufor2023}, and Mesorch~\cite{zhu2025mesorch} are evaluated in their original one-shot prediction paradigm. All methods are compared on both source-aligned (CASIA v1~\cite{CASIA_2013}) and cross-source datasets (Coverage~\cite{Coverage_2016}, Columbia~\cite{Columbia_2006}, NIST16~\cite{NIST16_2019}, IMD2020~\cite{IMD20_2020}, CocoGlide~\cite{trufor2023}, Autosplice~\cite{jia2023autosplice}), 
following the MVSS and the Cat-Net training protocol~\cite{ma2025imdl}.

As shown in Table~\ref{tab:performance}, our method consistently achieves strong performance across all datasets. Under both MVSS and CAT-Net protocols, it attains competitive source-aligned F1 scores while clearly outperforming prior methods on most cross-source benchmarks, leading to the best Cross-Source Avg and Overall Avg. \textbf{Notably}, most of our improvements come from the harder cross-source benchmarks, where existing models degrade but ours remain stable. This indicates that our model generalizes well rather than overfitting to source-specific cues, largely because the sequential prediction design encourages learning the manipulation process itself. Additional analyses are provided in the Supplementary Material.

\subsubsection{Ablation Study}
To evaluate the contribution of each component, we conduct ablations by selectively modifying or removing modules in our framework. As shown in Table~\ref{tab:ablation}, all variants lead to a performance drop compared with the full model.

\textbf{w/o TGF} replaces the proposed gated fusion with a plain decoder that directly consumes concatenated image and mask features. 
\textbf{w/o Edge Supervision} and \textbf{w/o Step Embedding} remove boundary guidance and temporal conditioning, respectively, each causing a slight drop. 
\textbf{w/o Multi-Scale} uses only the deepest-scale features, while \textbf{w/o Mask Feature} further removes the mask branch entirely and decodes solely from image features. 
The drops of these variants highlight the importance of mask-aware representation and multi-scale fusion.

We additionally test a \textbf{w/o Multi-Step } variant, in which the model directly outputs the final manipulation mask in one pass rather than generating stepwise binary updates autoregressively. This one-shot formulation produces a substantial drop, demonstrating that explicitly modeling the sequential editing process is critical for accurate localization. Finally, adding a DCT-based feature extractor (\textbf{w/ DCT Extractor}) further hurts performance, consistent with observations from IMDL-BenCo~\cite{ma2025imdl}.

\begin{table}[ht]
\centering
\caption{Ablation study}
\label{tab:ablation}
\renewcommand{\arraystretch}{1.1}
\begin{tabular}{lcc}
\toprule
\textbf{Variant} & \textbf{Average F1} & \textbf{Drop} \\
\midrule
\textbf{Full Model}         & \textbf{0.643} & -- \\
w/o TGF                     & 0.634 & \textcolor{gray}{-0.009} \\
w/o Edge Supervision        & 0.628 & \textcolor{gray}{-0.015} \\
w/o Step Embedding          & 0.618 & \textcolor{gray}{-0.025} \\
w/o Multi-Scale             & 0.572 & \textcolor{blue}{-0.071} \\
w/o Mask Feature            & 0.513 & \textcolor{blue}{-0.130} \\
w/o Multi-Step      & 0.432 & \textcolor{blue}{-0.211} \\
w/ DCT Extractor            & 0.402 & \textcolor{blue}{-0.241} \\
\bottomrule
\end{tabular}
\end{table}
\subsubsection{Model Complexity and Inference Efficiency}
We performed a comparative analysis across all models using a batch size of 1 and an input resolution of 512×512, evaluating parameter count, FLOPs, and inference time. As shown in Table~\ref{tab:param_flop}, our model requires only 55.567M parameters and 95.993 GFLOPs—substantially lower than recent approaches—and achieves competitive inference latency. Although PSCC-Net has the smallest parameter count, it incurs the highest computational cost (376.832 GFLOPs) and the slowest runtime, reflecting a trade-off between parameter efficiency and actual computational burden.

It is worth noting that the reported FLOPs refer to a single forward pass, whereas the inference time indicates the average latency to generate a single complete prediction.
\begin{table}[ht]
\centering
\caption{Comparison of model efficiency}
\label{tab:param_flop}
\renewcommand{\arraystretch}{1.1}
\resizebox{0.475\textwidth}{!}{%
\begin{tabular}{lccc}
\toprule
\textbf{Model} & \textbf{Parameters (M)} & \textbf{FLOPs (G)} & \textbf{Time (ms)} \\
\midrule
MVSS-Net   & 150.528 & 171.008 & 59.284 \\
PSCC-Net   & 3.668   & 376.832 & 103.155 \\
CAT-Net    & 116.736 & 136.216 & 97.687 \\
TruFor     & 68.697  & 236.544 & 71.526 \\
Mesorch    & 85.754  & 124.928 & 54.893 \\
Ours       & 55.567  & 95.993  & 93.216 \\
\bottomrule
\end{tabular}
}
\end{table}

\subsection{Sequence Manipulation Scenario}
\begin{table}[ht]
\centering
\caption{Effect of sampled path count on multi-step tampering.}
\label{tab:ablation_samplecount}
\renewcommand{\arraystretch}{1.1}
\begin{tabular}{cccc}
\toprule
\textbf{Sample Count} & \textbf{Score} & \textbf{F1\textsubscript{match}} & \textbf{\#Masks} \\
\midrule
1 & 0.455 & 0.463 & 16,161 \\
2 & \textbf{0.469} & \textbf{0.476} & \textbf{32,322} \\
3 & 0.462 & 0.470 & 48,483 \\
4 & 0.458 & 0.466 & 64,644 \\
\bottomrule
\end{tabular}
\end{table}
Using synthetic and HSIM data introduced in Section~\ref{sec:hsim}, we train and evaluate our method under multi-step manipulation scenarios. The synthetic dataset is derived from CASIAv2 through tree-structured copy-move augmentation, where each manipulation tree is decomposed into multiple valid manipulation paths. This process generates 3,680 composite images, with 16,161 autoregressive mask steps produced per sampled path, using a fixed random seed to ensure reproducibility.

For evaluation, we adopt the HSIM dataset. From HSIM, we derive 2,442 valid manipulation paths, which serve as the test set for evaluating autoregressive localization performance using the HSS metric.

\begin{table}[ht]
\centering
\caption{Results of AR variants with different backbones}
\label{tab:ablation_ar_backbones}
\renewcommand{\arraystretch}{1.1}
\begin{tabular}{lcc}
\toprule
\textbf{Model} & \textbf{Score} & \textbf{F1\textsubscript{match}} \\
\midrule
MambaVision  & 0.422 & 0.438 \\
Segformer        & 0.469 & 0.476 \\
ConvNeXt         & 0.452 & 0.463 \\
SwinTransformer  & 0.435 & 0.459 \\
ResNet           & 0.389 & 0.413 \\
\bottomrule
\end{tabular}
\end{table}
We conduct two types of experiments. First, we perform a controlled ablation~(Table \ref{tab:ablation_samplecount}) to assess the impact of varying the number of sampled training paths per manipulation tree. We find that performance improves with more samples, peaking at 2, after which over-sampling may introduce noise and redundancy. Second, we evaluated five different backbones with comparable parameter scales to power our RITA framework: SegFormer-B3~\cite{SegFormer_2021}, ConvNeXt-Base~\cite{Convnet_2022}, Swin Transformer-Base~\cite{Swin_2021}, ResNet-101~\cite{Resnet_2016}, and MambaVision-B~\cite{hatamizadeh2025mambavision}. The results are detailed in Table~\ref{tab:ablation_ar_backbones}. SegFormer-B3 achieves the best performance. Additional quantitative results are in the Supplementary Material.

\subsection{Case Study: Emulating Unseen Manipulations}

Beyond the above quantitative results, we demonstrate that RITA can directly guide process-centric data synthesis. Given a previously unseen forged image in the dataset, RITA decomposes the final manipulation into a plausible sequence of atomic operations (e.g., object erase $\rightarrow$ object insertion $\rightarrow$ self-blending) rather than treating it as a single opaque end mask.

The resulting edit sequence can then be reused as a template: by reapplying and slightly perturbing these atomic steps, we can synthesize new multi-step forgeries that preserve the manipulation style and editing logic of the original sample. In this way, RITA realizes the pipeline \emph{unseen sample $\rightarrow$ decomposed steps $\rightarrow$ synthetic surrogates $\rightarrow$ fine-tuned baseline $\rightarrow$ improved generalization}. Leveraging such step-wise programs to guide manipulation synthesis systematically enriches the training distribution and enhances the generalization of localization models. More details are provided in the supplementary material.

\section{Conclusion}
In this work, we introduce a paradigm shift in image manipulation localization, moving from one-shot prediction to ordered sequence prediction. We identify \textbf{dimensional collapse} in existing methods as a key limitation and propose \textbf{RITA}, the first autoregressive framework that breaks down manipulations layer-by-layer. Supported by our new \textbf{HSIM} dataset and \textbf{HSS} metric, our method effectively captures the sequential and structural nature of complex manipulation. 

Our experiments show that the sequential approach works better: RITA not only outperforms all current state-of-the-art methods on existing benchmarks, but also performs very well on hierarchical localization tasks and provides a solid baseline for future work. By explaining \textit{how} manipulations happen, our work offers deeper forensic insight beyond just \textit{where} manipulations occur, and as shown in our case study, it also enables process-centric emulation of unseen manipulations.



\section*{Ackonwledgement}
This work is supported by the National Natural Science Foundation of China, Young Scientists Fund (C Class)(No.62506251) and Sichuan Province Major Special Project (2024ZDZX0001-3).

{
    \small
    \bibliographystyle{ieeenat_fullname}
    \bibliography{main}
}

\clearpage
\section{Appendix}

\subsection{Current Status of Manipulation Datasets}

Most existing IML benchmarks are annotated in a one-shot manner, providing only the final binary manipulation mask. However, the underlying manipulation workflows that give rise to these datasets are far from single-step. Early benchmarks such as Columbia (2006)~\cite{Columbia_2006} and CASIA~v1.0 (2013)~\cite{CASIA_2013} largely consist of pure copy--paste operations. CASIA~v2.0 (2013)~\cite{CASIA_2013} already introduced additional post-processing steps such as geometric transforms and boundary smoothing.

Subsequent datasets demonstrate increasingly multi-stage editing procedures. NIST16 (2019)~\cite{NIST16_2019} and IMD20 (2020)~\cite{IMD20_2020} were constructed through professional forensic pipelines that involve object removal, inpainting, color adjustment, and blending — inherently multi-step sequences. The most recent datasets, including COCO-GLID~\cite{trufor2023}and AutoSplice (2023)~\cite{jia2023autosplice}, are based on diffusion models whose generative dynamics are iterative by design, making their manipulation process intrinsically multi-step.

Thus, although these datasets appear as one-shot benchmarks in their released annotations, the manipulations they contain are products of multi-stage generation processes. This structural inconsistency further motivates our sequence-prediction perspective, which aligns directly with the progressive nature of real-world manipulations.

\subsection{Data Distribution}

\subsection{Traditional IML Training Protocols}
Following the standard settings in IMDL-BenCo~\cite{ma2025imdl}, our experiments in the Traditional one-shot IML scenario adopt both the MVSS and CAT protocols, consistent with Section~4.3 of the main paper. The MVSS protocol uses CASIA~v2.0~\cite{CASIA_2013} as the sole training set and evaluates on CASIA~v1.0~\cite{CASIA_2013}, Columbia~\cite{Columbia_2006}, NIST16~\cite{NIST16_2019}, IMD20~\cite{IMD20_2020}, COCO-GLID~\cite{trufor2023}, and AutoSplice~\cite{jia2023autosplice}, providing a benchmark for cross-source and cross-manipulation generalization. The CAT protocol constructs a balanced multi-source training set from CASIA~v2.0, FantasticReality~v1~\cite{Fantastic_Reality}, IMD20, and the four tampCOCO~\cite{CAT-Net2022} subsets (sp/cm/bcm/bcmc), sampling 1840 images from each training subset per epoch~\cite{ma2025imdl} to ensure distributional consistency. Its evaluation set matches the MVSS protocol but excludes IMD20, enabling a fair assessment of unified learning under heterogeneous training distributions.

\subsection{Synthetic Multi-Step Dataset}
To support the proposed sequence prediction paradigm (Section~3.1.1 of the main paper), we construct a synthetic multi-step dataset based on the manipulated images of CASIA~v2.0 using our tree-structured reverse sampling algorithm. The resulting dataset contains 3680 samples, each exhibiting a multi-step editing trajectory with highly non-uniform step counts. The statistical distribution, visualized in Figure~\ref{fig:step_distribution}, reveals a long-tailed structure that closely reflects realistic manipulation processes. This dataset compensates for the limitations of existing IML benchmarks, which provide only the final manipulation mask and lack supervision over intermediate editing stages, thereby mitigating the dimensional collapse phenomenon associated with one-shot learning.

\begin{figure}[t]
    \centering
    \includegraphics[width=0.975\linewidth]{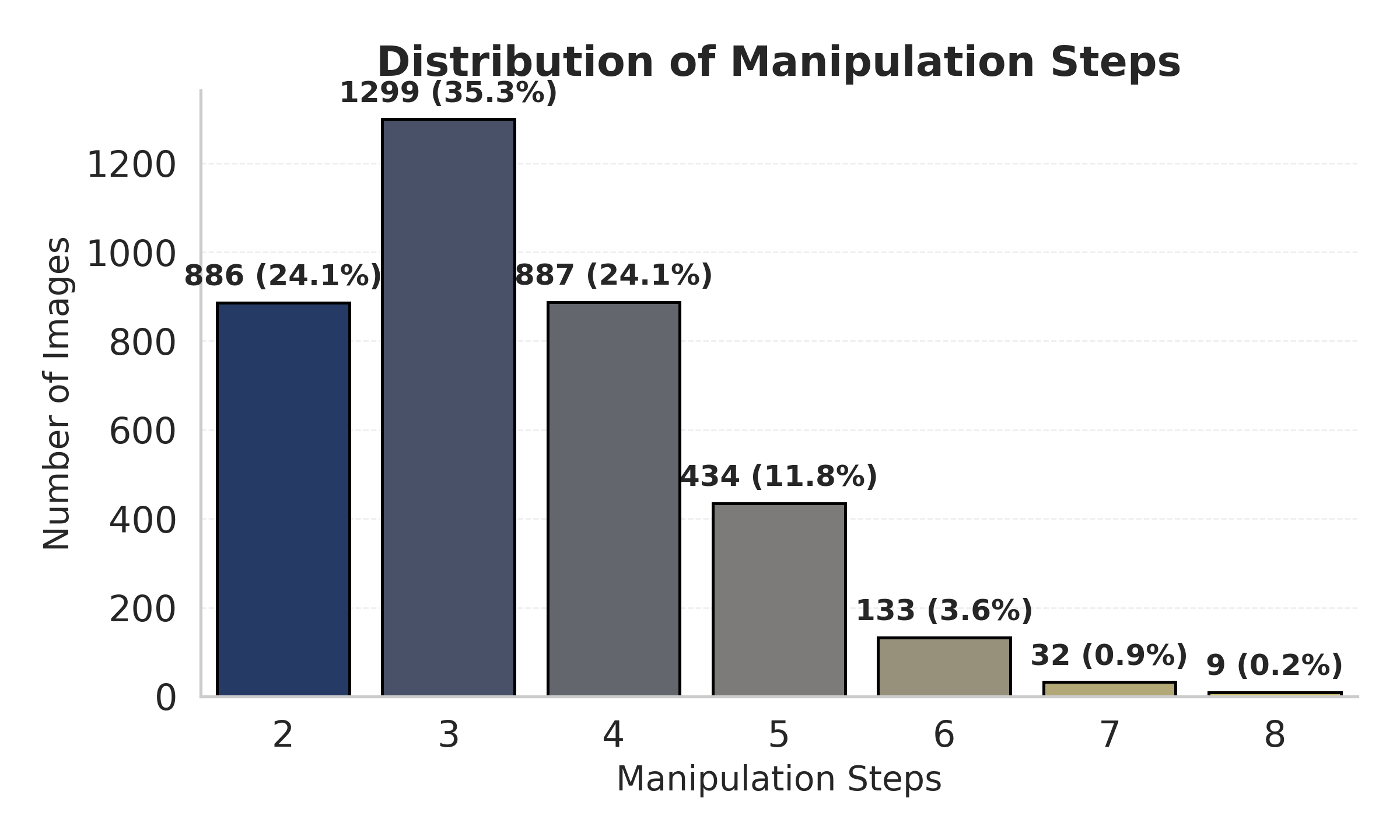}
    \caption{Distribution of manipulation step counts in the synthetic multi-step dataset.}
    \label{fig:step_distribution}
\end{figure}

\subsection{Hierarchical Sequence Structure of the Test Set}
The HSIM dataset serves as the test set in our sequence prediction experiments. Each image is constructed through a hierarchical and stepwise editing workflow: manipulation paths are first designed using GPT-4o and then realized via a progressive pixel-level refinement process assisted by GPT-Image-1. This results in multi-path samples with explicit temporal ordering and hierarchical dependency structures, fully aligned with the sequence prediction formulation in Section~4.4 of the main paper.

Two distributional properties of HSIM are particularly relevant to evaluating sequence-level generalization. First, the number of complete manipulation paths per image exhibits a clear long-tailed structure: 79\% of the images contain 1–10 valid paths, 19\% contain 10–20 paths, and only 2\% exceed 20 paths. This reflects realistic variability in manipulation-path complexity. Second, despite this heterogeneity in path counts, the editing depth within each path is highly consistent: every valid path in HSIM contains exactly 4–5 manipulation steps. This combination of ``diverse path counts but compact step depth'' makes HSIM a controlled yet representative test environment, enabling a precise assessment of the model’s cross-depth and cross-path generalization capabilities and its ability to capture the underlying temporal dependencies of manipulation sequences.

\begin{algorithm}[]
\caption{MonotonicF1Match$(P, M)$}
\label{alg:monotonic_match}
\KwIn{Predicted sequence $P = \{P_1, \dots, P_{T_p}\}$}
{Ground-truth sequence $M = \{M_1, \dots, M_{T_g}\}$}

\KwOut{Average F1 score of the optimal monotonic alignment}

\tcp{1. Compute pairwise similarity matrix}
Initialize matrix $F \in \mathbb{R}^{T_p \times T_g}$\;
\For{$i=1,\dots,T_p$}{
    \For{$j=1,\dots,T_g$}{
        $F[i, j] \leftarrow \text{F1\_score}(P_i, M_j)$\;
    }
}

\tcp{2. Find the optimal cumulative score via Dynamic Programming}
Initialize DP table $D \in \mathbb{R}^{T_p \times T_g}$ to store maximum cumulative scores\;
\For{$i=1,\dots,T_p$}{
    \For{$j=1,\dots,T_g$}{
        \If{$i=1$}{
            $D[i, j] \leftarrow F[i, j]$ \tcp*{Base case: first predicted step}
        } \Else{
            \tcp{Find max score from any valid previous alignment}
            $max\_prev\_score \leftarrow \max_{1 \le k \le j} D[i-1, k]$\;
            $D[i, j] \leftarrow F[i, j] + max\_prev\_score$\;
        }
    }
}

\tcp{3. Extract and normalize the final score}
$max\_cumulative\_score \leftarrow \max_{1 \le j \le T_g} D[T_p, j]$ \tcp*{Find the best path's total score}
\If{$T_p > 0$}{
    $F1_{match} \leftarrow \text{max\_cumulative\_score} / T_p$\;
} \Else{
    $F1_{match} \leftarrow 0$\;
}
\Return $F1_{match}$;
\end{algorithm}

\begin{table*}[ht]
\centering
\caption{Fair comparison between multi-step and single-step training methods.
All models use the same Synthetic Multi-Step Dataset; RITA is trained on intermediate masks, while baselines are trained only on the final mask. Results are evaluated under the MVSS protocol. Column-wise best scores are in \textcolor{red}{red}, second-best results are \underline{underlined}.}
\label{tab:multistep_performance}
\resizebox{\textwidth}{!}{%
\begin{tabular}{lccccccccc}
\toprule
\multirow{2}{*}{\textbf{Model}} & \multicolumn{1}{c}{\textbf{Source-Aligned}} & \multicolumn{7}{c}{\textbf{Cross-Source}} & \multirow{2}{*}{\textbf{Overall Avg}} \\
\cmidrule(lr){2-2} \cmidrule(lr){3-9}
& CASIAv1 & Coverage & Columbia & NIST16 & IMD2020 & CocoGlide & Autosplice & \textbf{Cross-Source Avg} & \\
\midrule
MVSS   & 0.534 & 0.259 & 0.386 & 0.246 & 0.279 & 0.291 & 0.294 & 0.292 & 0.327 \\
CAT-Net & \underline{0.581} & 0.296 & 0.584 & 0.269 & 0.273 & 0.290 & 0.354 & 0.344 & 0.378 \\
PSCC   & 0.381 & 0.286 & 0.573 & 0.185 & 0.251 & \underline{0.399} & \underline{0.487} & 0.363 & 0.366 \\
Trufor & 0.434 & \underline{0.331} & 0.689 & 0.257 & 0.258 & \underline{0.399} & 0.439 & 0.395 & 0.401 \\
Mesorch & \textcolor{red}{0.639} & 0.319 & \underline{0.744} & \textcolor{red}{0.321} & \underline{0.350} & 0.317 & 0.356 & \underline{0.401} & \underline{0.435} \\
Ours   & 0.422 & \textcolor{red}{0.357} & \textcolor{red}{0.793} & 0.306 & \textcolor{red}{0.360} & \textcolor{red}{0.503} & \textcolor{red}{0.661} & \textcolor{red}{0.497} & \textcolor{red}{0.486} \\
\bottomrule
\end{tabular}%
}
\end{table*}

\subsection{Evaluation on the Synthetic Multi-Step Dataset}
\subsection{Quantitative Experiments}
This experiment aims to verify whether RITA can still maintain leading performance when trained on the Synthetic Multi-Step Dataset, a dataset that contains multi-step manipulation trajectories. To ensure a fair comparison, all models are trained and evaluated under identical data settings, with the only difference lying in the form of supervision they receive.

RITA is trained with full multi-step supervision: the model learns to predict each intermediate manipulation mask in sequence and performs multi-step rollout during inference. The final-step prediction is treated as the final detection result, while any pixel predicted as manipulated at any intermediate step is unified as 1 to match the binary evaluation format used by baseline models.

Baseline models (MVSS~\cite{MVSS_2021}, CAT-Net~\cite{CAT-Net2022}, PSCC~\cite{liu2022pscc}, TruFor~\cite{trufor2023}, Mesorch~\cite{zhu2025mesorch}, etc.) are trained on the same images but can only access the final binary mask (single-step final mask) and do not observe any intermediate editing states. As shown in Table~\ref{tab:multistep_performance}, RITA consistently outperforms all single-step models across all six cross-source datasets in the MVSS protocol, achieving the highest overall and cross-source averages. Moreover, when we remove multi-step supervision and constrain RITA to single-step prediction (w/o multi-step), its Overall Avg drops from 0.486 to approximately 0.298, a decrease of nearly 20 percentage points. This further highlights the importance of maintaining the full sequence prediction framework in RITA.

\begin{figure}[ht]
  \centering
  \includegraphics[width=\linewidth]{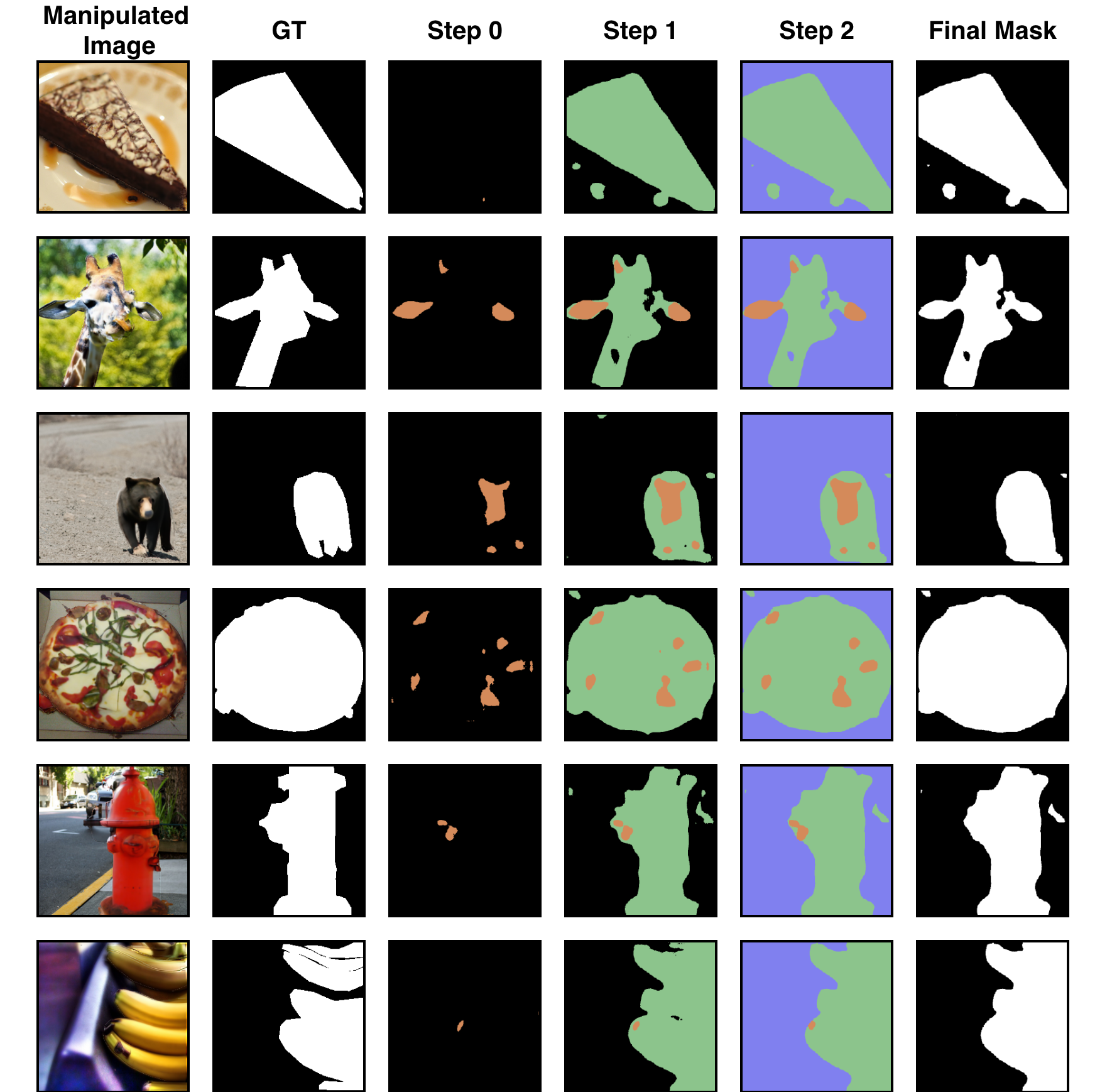}
\caption{Qualitative results of RITA on a one-shot test dataset. Although trained on a multi-step dataset, the model still exhibits a multi-stage refinement process: early steps emphasize salient manipulation cues, while later steps progressively correct errors and consolidate true manipulated regions, resulting in clean and coherent final masks.}
  \label{fig:ar2one}
\end{figure}

\subsection{Qualitative Experiments}
We further conduct a qualitative analysis to examine how RITA behaves when trained on multi-step synthetic data but evaluated on a standard one-step test set. As shown in Figure~\ref{fig:ar2one}, the model does not collapse into a single-shot predictor; instead, it retains a clear internal multi-stage reasoning process throughout inference. We consistently observe two characteristic behaviors: \textbf{(1) Salient-Cue First Activation}: in the first step, the model preferentially highlights the most prominent manipulation cues—such as boundary inconsistencies, structural discontinuities, or conspicuous texture anomalies—thus providing a coarse but meaningful initialization of manipulated regions; \textbf{(2) Progressive Error-Correction Refinement}: in the following steps, the model iteratively expands, suppresses, or reshapes these initial activations, correcting false positives and consolidating fragmented predictions into coherent manipulated regions. 

Across the entire test dataset, RITA follows a stable three-step prediction schedule, and this autoregressive pattern is consistently observed: the first step identifies salient manipulation traces, the second step further discovers additional related regions by conditioning on earlier predictions, and the third step revises and corrects errors introduced in previous steps to produce a clean and coherent final mask. These observations demonstrate that even in a one-step evaluation dataset

\subsection{Structure Matching with Dynamic Programming}

\label{appe:smdp}
This section provides the full algorithmic details of MonotonicMatch(Algorithm ~\ref{alg:monotonic_match}), which is referenced in the main paper (Sec. 3.1.3, Hierarchical Sequential Score) and deferred to the Supplementary Material.

A core challenge in evaluating the predicted path is that its length ($T_p$) may differ from the ground-truth path's length ($T_g$). A simple frame-by-frame comparison is therefore inadequate. To address this, we introduce \textbf{MonotonicMatch}, a dynamic programming algorithm designed to find the optimal alignment between the two sequences.

The goal of MonotonicMatch is to identify a monotonic, non-decreasing mapping between the steps of the predicted sequence and the ground-truth sequence. This mapping is “optimal” in that it maximizes the cumulative stepwise F1 score along the alignment path. First, we compute a pairwise F1 score matrix $F \in \mathbb{R}^{T_p \times T_g}$, where each element $F[i, j]$ represents the F1 score between the $i$-th predicted mask and the $j$-th ground-truth mask. Then, a dynamic programming table is populated to find the path that yields the highest average F1 score, enforcing the monotonic constraint. The detailed procedure is outlined in Algorithm \ref{alg:monotonic_match}.

\subsection{Robustness}
\begin{table*}[ht]
\centering
\caption{\textbf{Robustness under common image perturbations.}
  Entries are \emph{mean Binary F1} on the test set computed under the CAT{-}Net evaluation protocol.
  Perturbations include Gaussian noise (standard deviation), Gaussian blur (kernel size), and JPEG compression (quality factor).
  The rightmost \emph{Average} is the arithmetic mean of the per–condition means within each block (including ``None'').}

\label{tab:robustness}
\begin{tabular}{@{}clcccccccc@{}}
\toprule
\multirow{2}{*}{Pertubation}     & \multicolumn{1}{c}{\multirow{2}{*}{Model}} & \multicolumn{7}{c}{Sandard Deviations}                & \multirow{2}{*}{Average} \\ \cmidrule(lr){3-9}
                            & \multicolumn{1}{c}{} & None  & 3     & 7     & 11    & 15    & 19    & 23    &       \\ \midrule
\multirow{6}{*}{GaussNoise} & MVSS                 & 0.495 & 0.502 & 0.500 & 0.492 & 0.493 & 0.489 & 0.489 & 0.494 \\
                            & CAT-Net              & 0.533 & 0.512 & 0.500 & 0.484 & 0.473 & 0.462 & 0.454 & 0.488 \\
                            & PSCC                 & 0.555 & 0.539 & 0.531 & 0.522 & 0.521 & 0.518 & 0.512 & 0.528 \\
                            & Trufor               & 0.531 & 0.450 & 0.418 & 0.398 & 0.381 & 0.366 & 0.372 & 0.417 \\
                            & Mesorch              & 0.593 & 0.563 & 0.543 & 0.529 & 0.521 & 0.517 & 0.507 & 0.539 \\
                            & Ours                 & 0.643 & 0.622 & 0.609 & 0.602 & 0.598 & 0.592 & 0.590 & 0.608 \\ \midrule

\multirow{2}{*}{}                & \multicolumn{1}{c}{\multirow{2}{*}{}}      & \multicolumn{7}{c}{Kernel Size}                       & \multirow{2}{*}{Average} \\ \cmidrule(lr){3-9}
                            & \multicolumn{1}{c}{} & None  & 3     & 7     & 11    & 15    & 19    & 23    &       \\ \midrule
\multirow{6}{*}{GaussBlur}  & MVSS                 & 0.495 & 0.422 & 0.349 & 0.310 & 0.273 & 0.244 & 0.225 & 0.331 \\
                            & CAT-Net              & 0.533 & 0.487 & 0.458 & 0.429 & 0.417 & 0.402 & 0.392 & 0.445 \\
                            & PSCC                 & 0.555 & 0.509 & 0.454 & 0.414 & 0.377 & 0.343 & 0.310 & 0.423 \\
                            & Trufor               & 0.531 & 0.422 & 0.367 & 0.317 & 0.254 & 0.191 & 0.147 & 0.318 \\
                            & Mesorch              & 0.593 & 0.526 & 0.471 & 0.430 & 0.387 & 0.340 & 0.292 & 0.434 \\
                            & Ours                 & 0.643 & 0.577 & 0.523 & 0.505 & 0.485 & 0.467 & 0.456 & 0.508 \\ \midrule

\multirow{2}{*}{}                & \multicolumn{1}{c}{\multirow{2}{*}{}}      & \multicolumn{7}{c}{Quality Factors}                   & \multirow{2}{*}{Average} \\ \cmidrule(lr){3-9}
                            & \multicolumn{1}{c}{} & None  & 100   & 90    & 80    & 70    & 60    & 50    &       \\ \midrule
\multirow{6}{*}{JpegCompression} 
                            & MVSS                 & 0.495 & 0.493 & 0.462 & 0.434 & 0.408 & 0.392 & 0.369 & 0.436 \\
                            & CAT-Net              & 0.533 & 0.547 & 0.522 & 0.495 & 0.484 & 0.480 & 0.474 & 0.505 \\
                            & PSCC                 & 0.555 & 0.534 & 0.478 & 0.449 & 0.435 & 0.423 & 0.392 & 0.467 \\
                            & Trufor               & 0.531 & 0.481 & 0.437 & 0.408 & 0.387 & 0.366 & 0.318 & 0.418 \\
                            & Mesorch              & 0.593 & 0.577 & 0.527 & 0.508 & 0.506 & 0.495 & 0.465 & 0.524 \\
                            & Ours                 & 0.643 & 0.639 & 0.604 & 0.586 & 0.575 & 0.565 & 0.539 & 0.593 \\ 
\bottomrule
\end{tabular}
\end{table*}

\label{sec:robustness}
This robustness analysis belongs to the main paper, Section 4.3.1 (Performance Comparison), and complements the quantitative comparison by evaluating model stability under realistic perturbations.

To comprehensively assess the stability of different models under realistic conditions, we consider three representative perturbation families that widely occur in practical imaging scenarios: (i) \emph{Gaussian noise}, which emulates sensor-level corruption and thermal noise during acquisition; (ii) \emph{Gaussian blur}, which mimics defocus and motion blur frequently introduced by imperfect optics or camera shake; and (iii) \emph{JPEG compression}, which reflects storage and transmission artifacts due to lossy coding. 
For each perturbation family, we progressively increase the perturbation intensity, thereby creating a spectrum of degradation levels that range from mild to severe.

Following the CAT{-}Net evaluation protocol, we compute the mean Binary F1 score for every condition, and summarize the overall performance with a block-wise \emph{Average}, defined as the arithmetic mean of the per–condition means (including the ``None'' case). 
This evaluation protocol ensures fairness across methods and allows us to capture the sensitivity of each model to different perturbation strengths.

Results are reported in Table~\ref{tab:robustness}. 
Across all perturbation families, our method consistently achieves the highest averages, obtaining \textbf{0.608} under Gaussian noise, \textbf{0.522} under Gaussian blur, and \textbf{0.593} under JPEG compression. 
These results represent substantial margins over the strongest baselines (e.g., Mesorch: 0.539/0.434/0.524; CAT-Net: 0.488/0.445/0.505), highlighting the robustness of our approach. 
More importantly, we observe that prior methods degrade significantly when perturbation severity increases—for example, Trufor exhibits rapid performance drops under blur and noise, and PSCC becomes unstable under low-quality JPEG factors. 
By contrast, our method maintains comparatively stable scores even at the most extreme settings, such as large noise standard deviations, large blur kernels, and very low JPEG quality. 

Taken together, these experiments demonstrate that our model generalizes robustly to realistic degradations and is thus more reliable for deployment in unconstrained environments.

\subsection{Quantitative Experiment}
\subsection{Results on Traditional Datasets}
\label{sec:quantitative_tradition}
This qualitative analysis corresponds to the main paper, Section 4.3 (Existing Manipulation Scenarios), and complements the quantitative comparisons with visual evidence.

In addition to quantitative benchmarks, we further provide qualitative comparisons in Figure~\ref{fig:qualitative}. 
We randomly selected two non-semantically manipulated images and five semantically manipulated images according to their proportions in the dataset, covering diverse object categories, background contexts, and manipulation types. 
This setup ensures a representative evaluation across both manipulations with clear semantic meaning (e.g., replacing or altering salient objects) and manipulations that operate at a lower, often background or texture level without explicit semantic cues. 

Compared with state-of-the-art baselines, our method produces more precise and coherent masks. 
Specifically, the predicted regions not only align with the manipulated object layout but also preserve fine-grained boundaries, even in challenging cases involving subtle splicing, occlusion, or background-level editing. 
For semantically manipulated examples, our model accurately captures the global structure of manipulated objects while suppressing false positives in unaltered areas, which is crucial for practical scenarios where semantic consistency is essential. 
In contrast, existing baselines often either over-segment (producing large false-positive regions) or under-segment (missing critical manipulated parts), leading to incomplete or noisy masks.

For non-semantically manipulated cases, where manipulations are less visually salient and often manifest as texture inconsistencies or geometric misalignments, our autoregressive paradigm demonstrates robustness by sequentially refining predictions and yielding compact, accurate masks that isolate the true tampered regions. 
Unlike Traditional feed-forward architectures that may overlook weak or ambiguous traces, the autoregressive design enables iterative reasoning across spatial contexts, progressively consolidating local evidence into globally consistent predictions. 
Such a mechanism is particularly important in real-world images, where manipulations may be subtle and interwoven with natural variations.

Overall, these qualitative results confirm that our design generalizes well across manipulation types, successfully handling both semantically meaningful and background-level manipulations. 
They further illustrate that the autoregressive paradigm provides a principled way to enhance manipulation localization, leading to high-fidelity results and making our method better suited for deployment in diverse and unconstrained environments compared with prior approaches.

\begin{figure}[t]
  \centering
  \includegraphics[width=\linewidth]{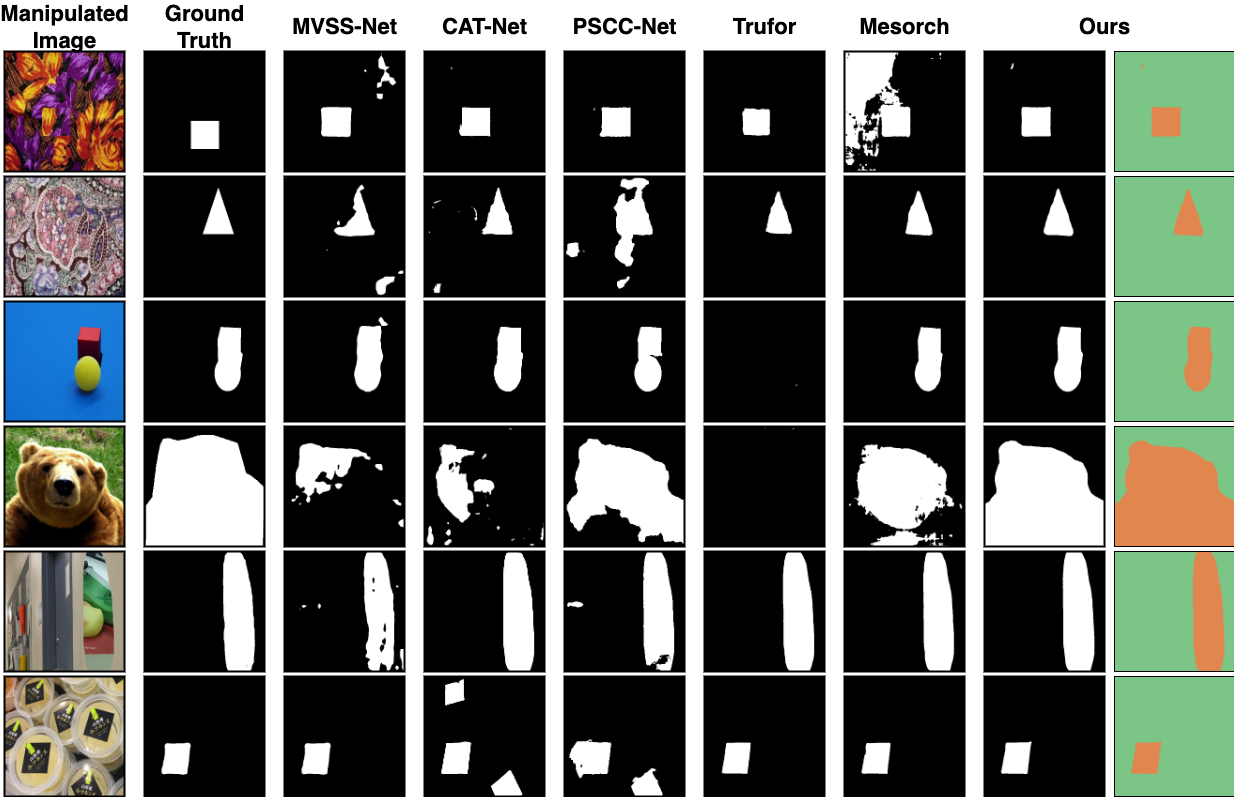}
  \caption{Qualitative analysis of SOTA models on Traditional datasets. 
We randomly selected and compared two semantically manipulated images and five non-semantically manipulated images according to their proportions in the dataset. 
The first two rows show non-semantically manipulated examples, while the last four rows correspond to semantically manipulated cases.  
The rightmost column presents our two-step reasoning results: the orange region indicates Step 1, and the green region Step 2.  
The second-to-rightmost column shows the corresponding 0/1 mask outputs.}
  \label{fig:qualitative}
\end{figure}

\begin{figure}[ht]
  \centering
  \includegraphics[width=\linewidth]{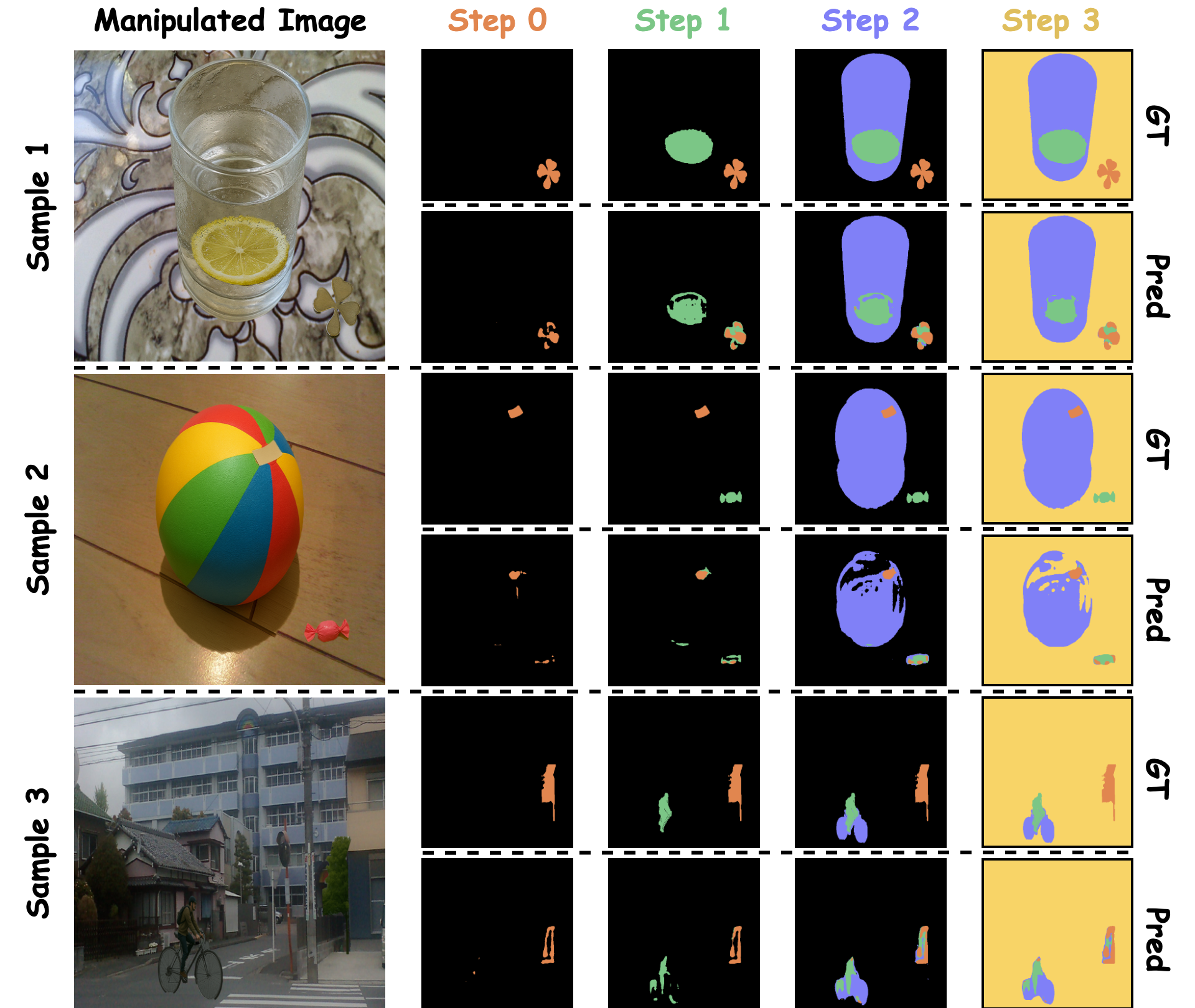}
  \caption{Qualitative results on the proposed HSIM dataset. 
  Each row corresponds to one sample, where columns show sequential tampering steps (Step 0–Step 3). 
  GT denotes the ground-truth mask at each stage, while Pred indicates our predictions. 
  Our method successfully tracks manipulation evolution across steps, aligning predictions with the progressive nature of multi-step edits.}
  \label{fig:multistep}
\end{figure}

\subsection{Results on Sequence Manipulation}

This qualitative analysis is part of the main paper, Section 4.4 (Sequence Manipulation Scenario), and provides visual illustrations of multi-step localization behavior.

\label{sec:quantitative_sequence}
To further evaluate the capability of our autoregressive framework, we conduct qualitative analysis on the proposed multi-step manipulation dataset, as shown in Figure~\ref{fig:multistep}. 
Unlike Traditional benchmarks where manipulations are applied once, these cases involve sequential tampering operations, progressively altering different regions or objects in the same image. 
Such a setup better reflects real-world scenarios, where images may undergo multiple edits over time.

As illustrated in Figure~\ref{fig:multistep}, our method is able to trace manipulation evolution across steps, from the initial local insertion (Step 0) to cumulative object-level alterations (Step 3). 
The predicted masks closely align with ground-truth annotations at each stage, successfully distinguishing newly introduced manipulations from previously existing edits. 
This progressive localization ability highlights the effectiveness of our autoregressive paradigm: rather than collapsing all manipulations into a single mask, it incrementally builds up tampering evidence in a temporally consistent manner. 

For example, in Sample~1, the small inserted pattern (Step 0) is correctly identified, followed by precise delineation of the lemon slice (Step 1) and cup boundary (Step 2). 
In Sample~2, where both object-level (ball) and small patch manipulations are present, our model captures the structural evolution while avoiding confusion between new and old tampering. 
Finally, in Sample~3, the method robustly localizes subtle manipulations across different semantic categories (e.g., structural edits on poles and bicycles), demonstrating generalization across diverse manipulation styles. 

These results suggest that our framework is inherently well-suited for multi-step tampering detection, offering interpretability by revealing \textit{how} manipulations accumulate and evolve, which is not possible with Traditional one-shot segmentation baselines.

\begin{figure}[ht]
  \centering
  \includegraphics[width=\linewidth]{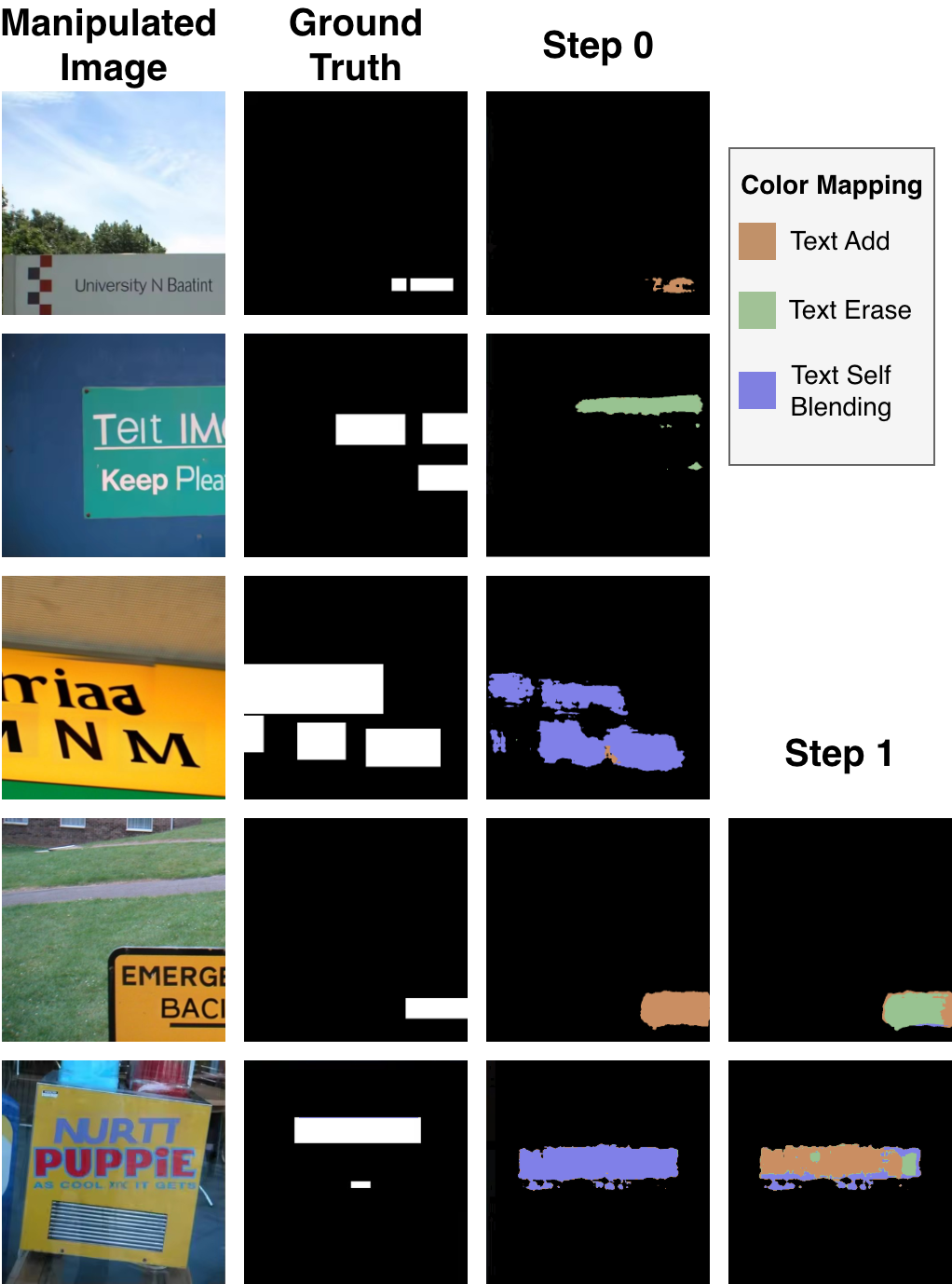}
  \caption{Examples of RITA-guided atomic-step decompositions on the OSTF dataset. RITA identifies meaningful atomic operations—such as text addition, text erasing, and text self-blending.}
  \label{fig:ostf}
\end{figure}

\subsection{RITA as a Process-Guided Data Synthesis Instructor}
This section provides the detailed content of the case study introduced in Section~4.5 of the main paper, illustrating how RITA guides the forgery synthesis pipeline to produce more precise and process-aligned synthetic data.

In manipulation localization, synthesizing training samples that include manipulation types unseen during training is one of the most important strategies for improving model generalization~\cite{park2025community,wang2025scaling}. Existing image and document forgery detectors typically rely on manually designed editing pipelines or randomly sampled combinations of atomic operations to create synthetic forgeries.~\cite{qu2023doctamper,qu2025ostf} These synthetic images are then fed to the base model in an attempt to improve robustness to newly emerging manipulation types. However, such heuristic strategies cannot guarantee realistic editing logic, and the generated manipulations often deviate significantly from the characteristics of these new forms of tampering.

In contrast, this case study demonstrates that RITA acts as a reliable and structured \emph{instructor} for data synthesis. Instead of relying on manually hypothesized operation sequences or randomly composed atomic operations, RITA extracts the latent manipulation process from an unseen sample, decomposes it into reusable atomic editing steps, and provides precise, process-aligned guidance on how synthetic data should be constructed. Importantly, RITA does \emph{not} directly synthesize images; rather, it provides target-domain editing programs that can be applied to clean images, enabling synthetic manipulations whose editing logic faithfully matches the target domain. This produces consistent, transferable, and stylistically accurate synthetic data, improving generalization to unseen manipulation types.

\subsection{Cross-Domain Setup and Baseline Performance}

We adopt an $A \rightarrow B$ cross-domain setting, where domain $A$ is TSROIE~\cite{wang2022tsroie} and domain $B$ is OSTF~\cite{qu2025ostf}. Three representative single-step document manipulation localization models (Mesorch~\cite{zhu2025mesorch}, FFDN~\cite{chen2024ffdn}, and DTD~\cite{qu2023doctamper}) are trained on the original TSROIE training set (2747 images) and evaluated on the OSTF test set to obtain baseline cross-domain performance.

We adopt an $A \rightarrow B$ cross-domain setting, where domain $A$ is TSOIRE~\cite{wang2022tsroie} and domain $B$ is OSTF~\cite{qu2025ostf}. These single-step models (Mesorch~\cite{zhu2025mesorch}, FFDN~\cite{chen2024ffdn}, and DTD~\cite{qu2023doctamper}) are trained using the ForensicHub~\cite{du2025forensichubunifiedbenchmark} framework on the original TSOIRE training set (2747 images) and then evaluated on the OSTF test set to obtain baseline cross-domain performance.

\subsection{Training RITA with Atomic Editing Operations}

We then train a RITA model on our private document dataset, which contains 23 strictly defined atomic manipulation types, including mosaic variants (such as \texttt{mosaic\_block\_blur} and \texttt{mosaic\_block\_random\_color}), text addition (\texttt{text\_add}), text erasing (\texttt{text\_erase}), text removal (\texttt{text\_remove}), and self-blending (\texttt{text\_self\_blending}). Unlike the RITA in the main paper, pixel values here represent atomic operation categories directly, and the temporal order is expressed only by the index of each step in the output sequence. Each predicted mask encodes one atomic operation type, and the full sequence forms a reverse-order editing program.

\subsection{Atomic Decomposition of Unseen OSTF Samples}

After training, we apply RITA to manipulated samples from OSTF.  
As shown in Fig.~\ref{fig:ostf}, RITA predicts a short and plausible atomic sequence, typically:
\[
\text{text\_erase} \rightarrow \text{text\_add} \rightarrow \text{text\_self\_blending}.
\]
Here, \texttt{text\_add} corresponds to inserting new text using the \texttt{PIL} rendering functions, \texttt{text\_self\_blending} is implemented by applying motion blur and median filtering to simulate blended text regions, and \texttt{text\_erase} is performed by using LaMa~\cite{suvorov2022resolution} to remove the specified text region.

\subsection{Constructing the Cross-Domain Augmented Set \texorpdfstring{TSROIE\_AUG}{TSROIE\_AUG}}

Using the atomic editing programs recovered from OSTF (domain $B$), we apply these programs to clean images from TSROIE (domain $A$) and construct 2551 guided synthetic samples, denoted as \texttt{TSROIE\_AUG}. RITA only provides the editing instructions; the actual synthesis is performed by applying these instructions to clean images. This aligns TSROIE's manipulation logic with OSTF while preserving its visual appearance, achieving $A \rightarrow B$ manipulation-style transfer.

We retrain Mesorch, FFDN, and DTD on the combined training set (TSROIE\_TRAIN + TSROIE\_AUG) and evaluate them on OSTF. As shown in Table~\ref{tab:guided}, all models achieve substantial improvements.

\begin{table}[h]
\centering
\caption{Effect of RITA-based atomic-program augmentation on the OSFT dataset. Random augmentation is denoted by $\dagger$.}
\label{tab:guided}
\begin{tabular}{lccc}
\toprule
\textbf{Model} & \textbf{Before} & \textbf{Afte} & $\Delta$ \\
\midrule
Mesorch & 0.1749 & \textbf{0.3770} & +0.2021 \\
{\footnotesize Mesorch\,$\dagger$ (random)} & --- & 0.0265 & --- \\
\midrule
FFDN    & 0.2665 & \textbf{0.3849} & +0.1184 \\
{\footnotesize FFDN\,$\dagger$ (random)} & --- & 0.0412 & --- \\
\midrule
DTD     & 0.1921 & \textbf{0.2311} & +0.0390 \\
{\footnotesize DTD\,$\dagger$ (random)} & --- & 0.0583 & --- \\
\bottomrule
\end{tabular}
\end{table}

\subsection{Ablation Study: Failure of Random Atomic Combinations}

To rule out the possibility that the performance gains arise from generic random augmentation, we perform a controlled ablation in which both the atomic operation types and their ordering are sampled uniformly at random. This form of augmentation provides no structural guidance, ignores the editing logic commonly present in real forgeries, and produces manipulation patterns that are highly inconsistent with those observed in the target domain.

As shown in Table~\ref{tab:guided}, random augmentation leads to extremely poor performance across all models. For example, the Mesorch model, which achieves 0.1749 before augmentation and 0.3770 with RITA-guided augmentation, collapses to only 0.0265 under random augmentation, losing almost all discriminative ability. Similar catastrophic degradation is observed for FFDN and DTD, confirming that random atomic combinations introduce severe distributional mismatches rather than useful diversity.

This stark contrast highlights that only structured, process-aligned atomic sequences recovered by RITA can produce effective synthetic data. Random compositions fail to approximate realistic manipulation processes and therefore provide no benefit for model generalization. RITA’s guidance is thus not merely helpful but essential for generating transferable and domain-consistent training samples.


\subsection{Future Works}
The research scope of RITA is not limited to 2D natural image manipulation localization. Since this work reformulates traditional static manipulation detection into a progressively unfolded temporal modeling process at the methodological level, its core idea does not rely on a single 2D pixel-based representation but instead shows potential for transfer to broader visual content security tasks. In the future, RITA may not only continue to serve natural image scenarios but also be extended to more complex tasks, such as remote sensing image manipulation~\cite{11418730,10945380}, detection, and manipulation analysis for higher-dimensional digital humans~\cite{jia2026ram,li2025human,li2026multiple,gu2025mocount,li2025chatmotion} and 3D assets~\cite{11299106}. Compared with ordinary natural images, these forms of content usually exhibit stronger structural properties, temporal dependencies, and spatial hierarchies, while their manipulations are no longer limited to simple local texture editing, but may involve region replacement, structural perturbation, geometric deformation, motion reordering, and even semantically driven composite modifications. Therefore, starting from 2D natural images and further extending along both the content dimension and the representation dimension to domains such as remote sensing, digital humans, and 3D assets would not only help verify the generality of RITA, but also promote the development of manipulation detection from conventional image forensics toward a broader intelligent content security framework.

\subsection{Ethics Statement}
The HSIM dataset used in this paper was entirely created and manually annotated by the authors’ research team. No external annotators or third-party contributors were involved in the data collection, manipulation, or labeling process. All data were generated under controlled conditions, contain no personally identifiable information, and fully comply with ethical standards for academic research.

\end{document}